\documentclass[letterpaper, 10 pt, conference]{ieeeconf}  

\IEEEoverridecommandlockouts                              

\usepackage{graphics} 
\usepackage{epsfig} 
\usepackage{mathptmx} 
\usepackage{times} 
\usepackage{amsmath} 

\usepackage{gensymb}

\usepackage{cite}
\usepackage{textcomp}
\usepackage{amsmath,amssymb,amsfonts}
\usepackage{multicol}
\usepackage{float} 
\usepackage{tablefootnote}
\usepackage{url}
\usepackage{hyperref}
\usepackage{subfigure}
\usepackage{multirow}
\usepackage{array}
\usepackage{xcolor}
\newcolumntype{P}[1]{>{\centering\arraybackslash}p{#1}}

\usepackage{scalerel}
\usepackage{tikz}
\usetikzlibrary{svg.path}

\definecolor{orcidlogocol}{HTML}{A6CE39}
\tikzset{
  orcidlogo/.pic={
    \fill[orcidlogocol] svg{M256,128c0,70.7-57.3,128-128,128C57.3,256,0,198.7,0,128C0,57.3,57.3,0,128,0C198.7,0,256,57.3,256,128z};
    \fill[white] svg{M86.3,186.2H70.9V79.1h15.4v48.4V186.2z}
                 svg{M108.9,79.1h41.6c39.6,0,57,28.3,57,53.6c0,27.5-21.5,53.6-56.8,53.6h-41.8V79.1z M124.3,172.4h24.5c34.9,0,42.9-26.5,42.9-39.7c0-21.5-13.7-39.7-43.7-39.7h-23.7V172.4z}
                 svg{M88.7,56.8c0,5.5-4.5,10.1-10.1,10.1c-5.6,0-10.1-4.6-10.1-10.1c0-5.6,4.5-10.1,10.1-10.1C84.2,46.7,88.7,51.3,88.7,56.8z};
  }
}

\newcommand\orcidicon[1]{\href{https://orcid.org/#1}{\mbox{\scalerel*{
\begin{tikzpicture}[yscale=-1,transform shape]
\pic{orcidlogo};
\end{tikzpicture}
}{|}}}}

\usepackage{graphicx} 

\normalmarginpar 
\newcommand{\IEEEpreprintnotice}{
    \rotatebox{90}{
    \footnotesize
    © 2026 IEEE. This work has been submitted to the IEEE for possible publication.
    Copyright may be transferred without notice, after which this version may no longer be accessible.
    }
}

\title{\LARGE \bf
Nonlinear Performance Degradation of Vision-Based Teleoperation under Network Latency
}

\author{Aws Khalil$^{1}$\orcidicon{0000-0001-9139-3900} and Jaerock Kwon$^{2}$\orcidicon{0000-0002-5687-6998}
\thanks{Both authors are with the Department of Electrical and Computer Engineering, University of Michigan - Dearborn, Dearborn, MI, USA.
{\tt\small awskh@umich.edu$^{1}$},
{\tt\small jrkwon@umich.edu$^{2}$}}%
}

\begin{document}
    \maketitle
    \marginpar{\IEEEpreprintnotice} 

\begin{abstract}

Teleoperation is increasingly being adopted as a critical fallback for autonomous vehicles. However, the impact of network latency on vision-based, perception-driven control remains insufficiently studied. The present work investigates the nonlinear degradation of closed-loop stability in camera-based lane keeping under varying network delays. To conduct this study, we developed the Latency-Aware Vision Teleoperation testbed (LAVT), a research-oriented ROS~2 framework that enables precise, distributed one-way latency measurement and reproducible delay injection. Using LAVT, we performed 180 closed-loop experiments in simulation across diverse road geometries. Our findings reveal a sharp collapse in stability between 150 ms and 225 ms of one-way perception latency, where route completion rates drop from 100\% to below 50\% as oscillatory instability and phase-lag effects emerge. We further demonstrate that additional control-channel delay compounds these effects, significantly accelerating system failure even under constant visual latency. By combining this systematic empirical characterization with the LAVT testbed, this work provides quantitative insights into perception-driven instability and establishes a reproducible baseline for future latency-compensation and predictive control strategies.
Project page, supplementary video, and code are available at \url{https://bimilab.github.io/paper-LAVT}

\end{abstract}
    
    
    \section{Introduction}
\label{sec:introduction}

Teleoperation has emerged as an important fallback and transitional capability for autonomous driving systems, enabling remote intervention when onboard autonomy is unable to safely resolve a situation. Recent industrial deployments and research prototypes increasingly rely on teleoperation to handle edge cases, disengagements, and recovery scenarios without requiring a safety driver onboard \cite{kerbl2025tum, schimpe2022open}. As teleoperation becomes more tightly coupled with autonomous driving pipelines, understanding its operational limitations under realistic network conditions has become a critical research challenge.

A fundamental limitation of teleoperation is its sensitivity to network-induced latency. Teleoperated driving relies on a closed-loop exchange of sensory information and control commands between a remote client and a vehicle-side execution stack. Delays in this loop introduce stale observations and delayed actuation, which can degrade control performance and, in extreme cases, induce instability. This challenge is particularly pronounced in continuous control tasks such as lane keeping, where timely perception and feedback are essential for maintaining stable vehicle behavior.

Prior work has demonstrated that increasing network latency leads to measurable degradation in teleoperated driving performance, and has established latency tolerance benchmarks under controlled experimental conditions \cite{alsolami2025evaluating}. However, existing system-centric latency studies predominantly rely on localization- or map-based control pipelines, often using LiDAR perception and geometric path-following controllers. In such settings, perception is relatively decoupled from control timing, and sensing delay has a limited impact on control stability.

In contrast, vision-based teleoperation introduces a distinct set of challenges. Camera-driven perception pipelines operate directly on image data and are inherently sensitive to temporal misalignment between visual observations and vehicle motion. Even moderate perception delay can result in outdated lane geometry, shifted visual features, or misaligned control inputs, particularly in closed-loop control settings. Although cameras are widely used in autonomous driving and teleoperation systems, the impact of network latency on \emph{vision-based, perception-driven lane keeping} has not been systematically studied.

This paper addresses this gap by focusing explicitly on the effect of network-induced latency on camera-based lane keeping under teleoperation. Rather than proposing new teleoperation architectures, prediction mechanisms, or latency mitigation strategies, the goal of this work is to \emph{characterize} how delayed visual feedback affects closed-loop control behavior in a controlled and repeatable setting. To this end, we introduce a research-oriented ROS~2 teleoperation system, referred to as the \emph{Latency-Aware Vision Teleoperation testbed} (LAVT), designed specifically to support systematic latency studies for vision-based control.

LAVT enables distributed execution across client and server nodes, explicit time-stamping of perception and control signals, and controlled injection of network latency using standard Linux traffic control tools. While the system is deployable on a full-scale drive-by-wire research vehicle, all experiments in this paper are conducted in simulation to ensure safety, repeatability, and isolation of latency effects from confounding real-world factors. LAVT is designed specifically to enable controlled and reproducible latency characterization in vision-based teleoperation. The framework prioritizes measurement transparency, timing fidelity, and experimental controllability over feature completeness, making it suitable for systematic latency analysis across both simulation and real-vehicle deployments.

Using LAVT, we perform a systematic evaluation of how increasing network latency affects the stability and tracking performance of a vision-based lane-keeping controller operating in a closed loop. The study highlights degradation trends and failure modes that emerge as perception and control become increasingly desynchronized. These results provide empirical insights into the sensitivity of perception-driven control to latency and inform the design of future teleoperation and autonomous driving systems that rely on camera-based perception.

The contributions of this paper are threefold:
\begin{itemize}

    \item We provide empirical insights into degradation patterns and failure modes specific to perception-driven control under delayed visual feedback, without introducing mitigation or predictive strategies.
    \item We conduct a systematic simulation-based evaluation of how network-induced latency impacts stability and tracking performance in vision-based teleoperated lane keeping.
    \item We present LAVT, a research-oriented ROS~2-based teleoperation testbed designed to enable controlled studies of network latency effects on vision-based closed-loop driving, implemented in the CARLA simulation \cite{Dosovitskiy17} and integrated with a full-scale drive-by-wire research vehicle.    

\end{itemize}

    \section{Related Work}
\label{sec:related-work}

Teleoperation has been widely studied in autonomous driving, including system architectures, operator interfaces, communication pipelines, and human-in-the-loop evaluation. This section focuses on prior work most relevant to network-induced latency in teleoperated driving and vision-based control.

\subsubsection{Teleoperation Systems and Frameworks for Autonomous Driving}

Several teleoperation software stacks have been proposed to enable remote driving and assistance for autonomous vehicles. Open-source and modular frameworks support distributed deployment, operator interfaces, and integration with simulation and real vehicles \cite{schimpe2022open, hofbauer2020telecarla, gontscharow2025open}. Kerbl et al.\ introduced a ROS~2-based framework integrating remote driving, remote assistance, and Autoware, emphasizing modularity and deployability \cite{kerbl2025tum}.

These works primarily address architectural design and operational feasibility. Although they recognize latency as a system concern, they do not explicitly isolate or quantify the latency in closed-loop, perception-driven vehicle control. In contrast, the present study uses a dedicated testbed to systematically analyze latency-induced degradation in vision-based lane keeping.

\subsubsection{Latency in Teleoperated Driving}

The impact of network latency on teleoperated driving has been investigated in several studies. Alsolami et al.\ conducted a system-centric evaluation using LiDAR-based localization and geometric path-following control, identifying latency tolerance thresholds under controlled delay injection \cite{alsolami2025evaluating}. 

Other work has examined teleoperation reliability under network impairments, often emphasizing human operators. Tener et al.\ analyzed constrained-network teleoperation focusing on operator behavior and safety \cite{tener2022driving, tener2023design, tener2023toward, tener2024towards}. Implementation-oriented feasibility studies were reported by Kim et al.\ and Sato et al.\ \cite{kim2024teleoperated, kim2025development, sato2021implementation}. From a networking perspective, Fezeu et al.\ analyzed teleoperation over commercial 5G networks, highlighting tail-latency variability \cite{fezeu2025teleoperating}.

While these works demonstrate that latency affects teleoperated driving performance, they primarily consider human-in-the-loop systems or localization-based control pipelines. The degradation mechanisms introduced by delayed visual perception in camera-driven closed-loop control remain insufficiently studied.
\begin{figure*}[!t]
    \centering
    \includegraphics[width=\linewidth]{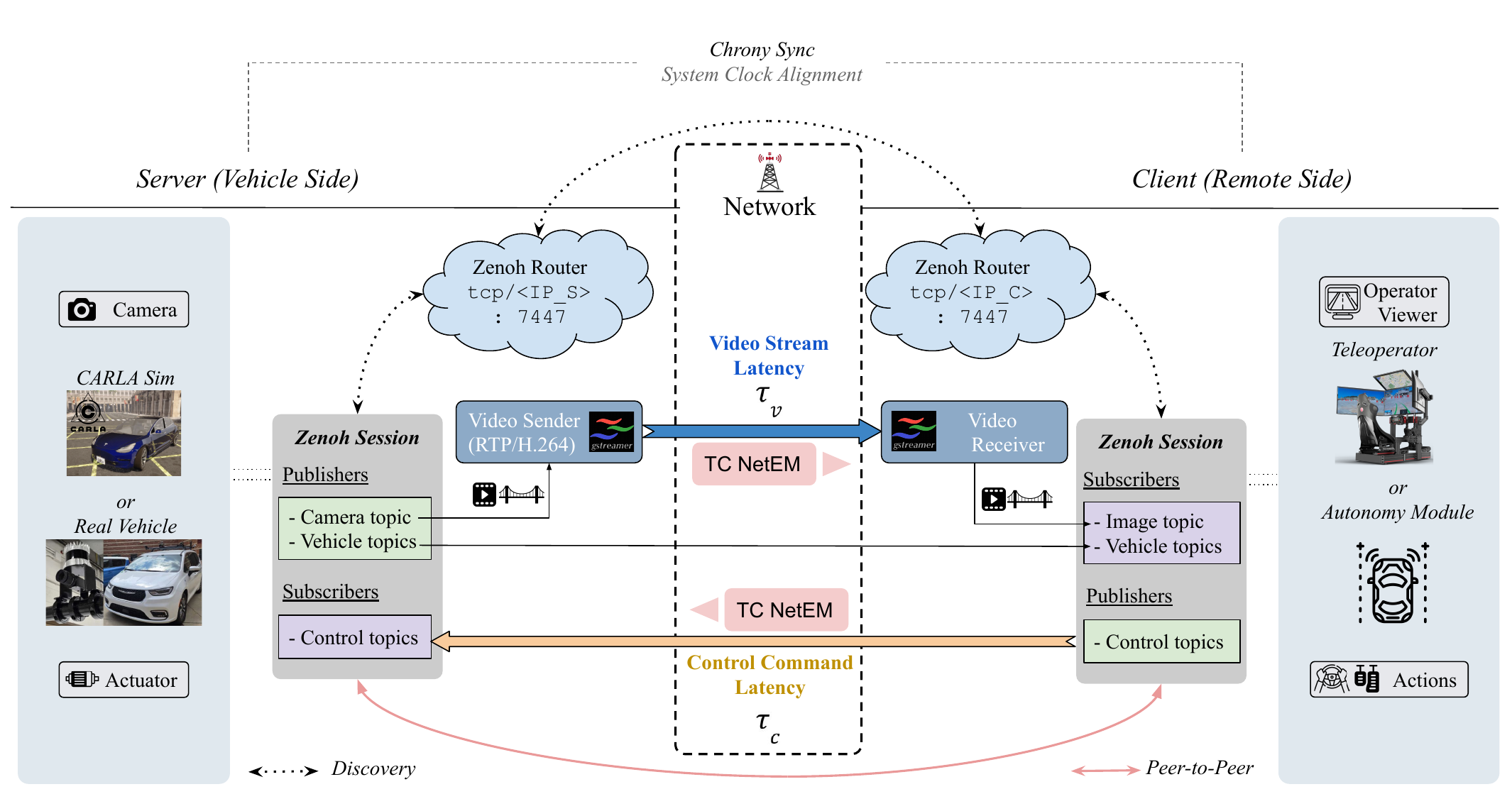}
    \caption{
    Architecture of the Latency-Aware Vision Teleoperation testbed (LAVT). The system operates in a distributed client–server configuration. On the vehicle side (server), a forward-facing camera publishes a local ROS~2 image topic. A video bridge node subscribes to this topic and transmits frames via a dedicated RTP/H.264 stream (GStreamer) over UDP. On the remote side (client), a corresponding video bridge decodes the stream and republishes the received frames as a local ROS~2 image topic for downstream processing by either a teleoperator interface or a vision-based autonomy module. Control commands are exchanged as ROS~2 topics over \texttt{rmw\_zenoh}, with independent Zenoh routers running on both server and client and connected via configured peer-to-peer sessions. The blue path denotes the perception channel with one-way video latency $\tau_v$, while the orange path denotes the actuation channel with one-way control latency $\tau_c$. Linux Traffic Control and Network Emulator (TC NetEm) is applied independently on each direction to inject controlled delay. Chrony provides system-level clock synchronization between machines, enabling accurate offset estimation and one-way latency computation using embedded frame timestamps. The resulting architecture forms a fully closed-loop teleoperation system suitable for systematic latency characterization in both simulation and real-vehicle deployments.
    }
    \label{fig:lavt-arch}
\end{figure*}

\subsubsection{Vision-Based Control and Perception Delay}

Vision-based perception is central to modern lane detection and lateral control, but camera pipelines are sensitive to temporal misalignment between perception and vehicle motion. Prior research has studied perception delay and mitigation strategies in autonomous driving and robotics \cite{tian2019motion, su2025mitigating}, and surveys have identified latency as a fundamental challenge in networked robotic systems \cite{kamtam2024network}.

In teleoperation, video streaming under network constraints has largely been evaluated in terms of operator situational awareness and communication efficiency \cite{stornig2021video, deng2025telesim}, without analyzing closed-loop autonomous control under delayed visual feedback.

The present work focuses on characterizing the impact of network-induced latency on a vision-based, perception-driven lane-keeping controller operating in a closed loop, exposing degradation patterns specific to delayed camera feedback.
    \section{Method}
\label{sec:method}

\subsection{Latency-Aware Vision Teleoperation Testbed}
\label{sec:method_overview}

LAVT is a distributed ROS~2 framework for controlled evaluation of network-induced latency in vision-based closed-loop teleoperation. The system separates perception, transport, and control layers while providing explicit time-stamping and one-way latency measurement.

LAVT operates across two nodes: a \emph{server} (vehicle side) and a \emph{client} (remote side).

\paragraph{Server}
The server hosts the vehicle platform (CARLA or a DBW-enabled vehicle), acquires camera frames, embeds send-time stamps, streams video via RTP/H.264 over UDP, receives control commands, and applies actuation.

\paragraph{Client}
The client receives and decodes the video stream, extracts embedded timestamps for one-way latency estimation, executes a vision-based lane-keeping controller, and transmits time-stamped control commands.

Both human teleoperation and remote autonomous driving are supported. All experiments in this study use a deterministic client-side controller to eliminate human reaction variability and ensure repeatable latency sensitivity analysis. The autonomy module shares the same communication pathways as teleoperation, so injected network delay affects both modes identically.

\begin{figure}[h]
\centering
\includegraphics[width=\linewidth]{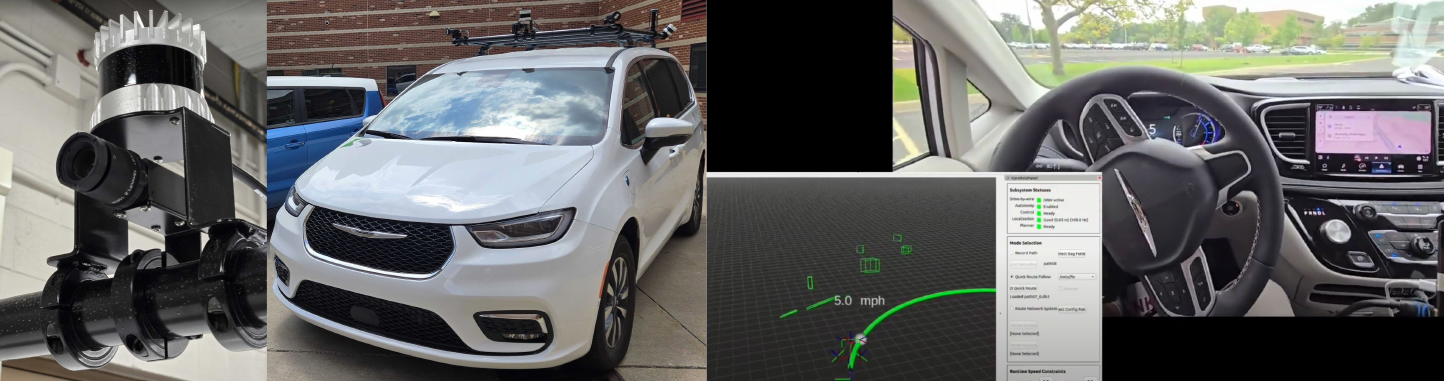} 
\caption{
A full-scale drive-by-wire (DBW) research vehicle integrated with LAVT for real-world deployment. The DBW system includes brake, throttle, steering, and shift-by-wire control modules. The platform is equipped with a multi-sensor suite including LiDAR, RGB cameras, and GPS. While quantitative latency experiments in this study are conducted in simulation, the complete teleoperation and streaming stack has been integrated with this vehicle without architectural modification.
}
\label{fig:jupiter}
\end{figure}

\paragraph{Real-vehicle integration}
Although quantitative experiments are conducted in simulation, LAVT has been integrated with a full-scale drive-by-wire research vehicle (Fig.~\ref{fig:jupiter}). Only sensing and actuation interfaces differ; streaming, synchronization, and latency measurement remain unchanged.

\subsection{ROS 2 Middleware Configuration}
\label{sec:method_middleware}

LAVT uses ROS~2 with the \texttt{rmw\_zenoh\_cpp} backend \cite{rmw_zenoh} to support routed and peer-to-peer deployment across distributed networks. This property is particularly relevant for teleoperation scenarios in which vehicle and operator nodes may be separated by routed networks or cloud infrastructure. Zenoh \cite{zenoh_project} enables flexible discovery and communication across heterogeneous network topologies without modifying application-layer logic.

\subsection{Closed-Loop Control Under Asymmetric Delay}
\label{sec:method_closed_loop}

We model the closed-loop teleoperation dynamics with separate perception and actuation delays. Let $I_t$ denote the camera observation captured on the server at time $t$, and let $u_t$ denote the control command applied to the vehicle at time $t$. The remote policy/controller $\pi(\cdot)$ operates on delayed observations,
\begin{equation}
u_t = \pi\!\left(I_{t-\tau_v}\right),
\label{eq:policy_delay}
\end{equation}
where $\tau_v$ is the end-to-end video (server$\rightarrow$client) latency. The vehicle state $x_t$ evolves under delayed actuation,
\begin{equation}
x_{t+1} = f\!\left(x_t, u_{t-\tau_c}\right),
\label{eq:actuation_delay}
\end{equation}
where $\tau_c$ is the control (client$\rightarrow$server) latency. This formulation cleanly separates perception and actuation delay channels and supports independent latency injection for each direction (Section~\ref{sec:method_injection}).

\subsection{Video Transport via GStreamer RTP/H.264}
\label{sec:method_gstreamer}

Video is transmitted using RTP over UDP with H.264 encoding via GStreamer \cite{gstreamer_project}. The server pipeline is:

\texttt{appsrc} $\rightarrow$ \texttt{H.264 encoder} $\rightarrow$ \texttt{rtph264pay} $\rightarrow$ \texttt{udpsink}

The client pipeline is:

\texttt{udpsrc} $\rightarrow$ \texttt{rtpjitterbuffer} $\rightarrow$ \texttt{rtph264depay} $\rightarrow$ \texttt{decoder} $\rightarrow$ \texttt{appsink}

RTP payloading, depayloading, and jitter buffering elements \cite{gstreamer_rtph264pay, gstreamer_rtph264depay, gstreamer_jitterbuffer} introduce codec and buffering delays depending on configuration. Measured latencies under each experimental condition are reported in Section~\ref{sec:results}.

\subsection{Time Synchronization and One-Way Latency Measurement}
\label{sec:method_timesync}

Accurate one-way latency measurement requires cross-machine clock alignment. All nodes run Chrony \cite{chrony_project, chrony_redhat} to discipline system clocks and minimize drift.

The inter-machine clock offset $\Delta$ is obtained directly from Chrony synchronization reports:

\begin{equation}
\Delta \triangleq t_{\text{server}} - t_{\text{client}}.
\label{eq:clock_offset}
\end{equation}

An optional RTT-based probe provides redundancy under a symmetric-delay assumption. The estimated $\Delta$ is used to convert timestamps when computing one-way video and control latency. Because latency injection is applied to the data plane, injected delay alters packet transit times without modifying underlying clock alignment.

\subsubsection{Video Send-Time Embedding}
\label{sec:method_stamp}

The server embeds a 64-bit nanosecond timestamp $t^{\text{server}}_{\text{send}}$ into each frame prior to H.264 encoding using an 8$\times$8 pixel stamp. The client extracts this value and computes one-way video latency:

\begin{equation}
\tau_v = t^{\text{client}}_{\text{recv}} - \left(t^{\text{server}}_{\text{send}} - \Delta\right).
\label{eq:video_latency}
\end{equation}

\subsubsection{Control Latency}
\label{sec:method_control_latency}

Control commands are time-stamped at transmission from the client. The server extracts the value and read one-way control latency computed as:

\begin{equation}
\tau_c = t^{\text{server}}_{\text{rx}} - \left(t^{\text{client}}_{\text{tx}} + \Delta\right).
\label{eq:control_latency}
\end{equation}

\begin{figure}[!t]
\centering
\includegraphics[width=\linewidth]{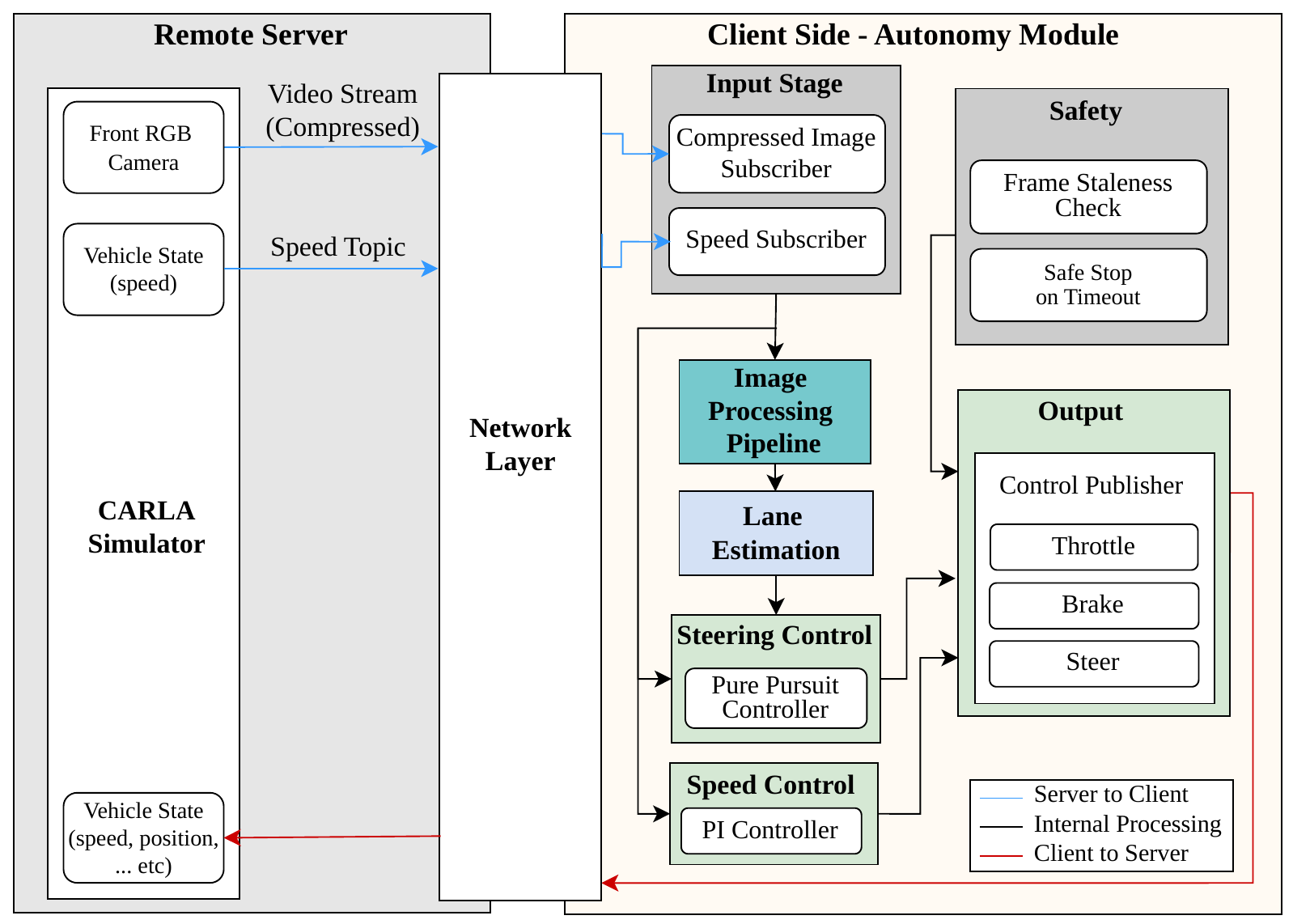}
\caption{
Client-side autonomy module in LAVT. The module receives a compressed video stream and vehicle speed from the server. A deterministic classical vision pipeline estimates lane boundaries in a bird’s-eye-view (BEV) representation, from which a centerline is computed. Lateral control is generated using a speed-adaptive Pure Pursuit controller, and longitudinal motion is regulated by a PI speed controller. Safety mechanisms include frame staleness detection and safe-stop behavior.
}
\label{fig:autonomy}
\end{figure}

\subsection{Latency Injection Model (NetEm)}
\label{sec:method_injection}

Network delay is injected using Linux Traffic Control (\texttt{tc}) with the \texttt{NetEm} queueing discipline \cite{netem_manpage}. Delay is applied independently on (i) the server egress interface (video) and (ii) the client egress interface (control), enabling separate manipulation of $\tau_v$ and $\tau_c$.

Baseline runs include only inherent transport and codec buffering. Experimental conditions apply additional constant delay (and optional jitter/loss) while measured latencies are verified using the framework in Section~\ref{sec:method_timesync}.

\subsection{Client-Side Autonomy Module}
\label{sec:method_autonomy}

Figure~\ref{fig:autonomy} illustrates the high-level architecture of the client-side autonomy module. It is a deterministic classical vision-based lane-keeping controller designed to expose latency sensitivity without learning-based compensation. The module operates exclusively on the received compressed image stream and vehicle speed topic, without access to simulator ground-truth.
A lightweight image-processing pipeline performs region-of-interest cropping, color/gradient thresholding, and bird’s-eye-view (BEV) transformation. Lane boundaries are estimated via sliding-window initialization followed by polynomial tracking. A lane confidence metric governs speed reduction and safety fallback.
Lateral control is implemented using a speed-adaptive Pure Pursuit controller \cite{coulter1992implementation} in the BEV domain, while longitudinal motion is regulated by a PI controller targeting constant speed. Safety mechanisms include frame staleness detection and controlled braking under persistent perception failure.
    \section{Experimental Design}
\label{sec:experimental-design}

This section describes the protocol used to quantify latency-induced degradation in closed-loop vision-based lane keeping using LAVT.

\subsection{Simulation Environment}
\label{sec:exp_environment}

All experiments are conducted in CARLA to ensure safety and repeatability. A single forward-facing RGB camera serves as the sole perception input. No ground-truth pose or simulator state is provided, ensuring that control relies exclusively on delayed visual feedback.

\subsection{Route Design}
\label{sec:exp_routes}

\begin{figure}[!t]
\centering
\includegraphics[width=\linewidth]{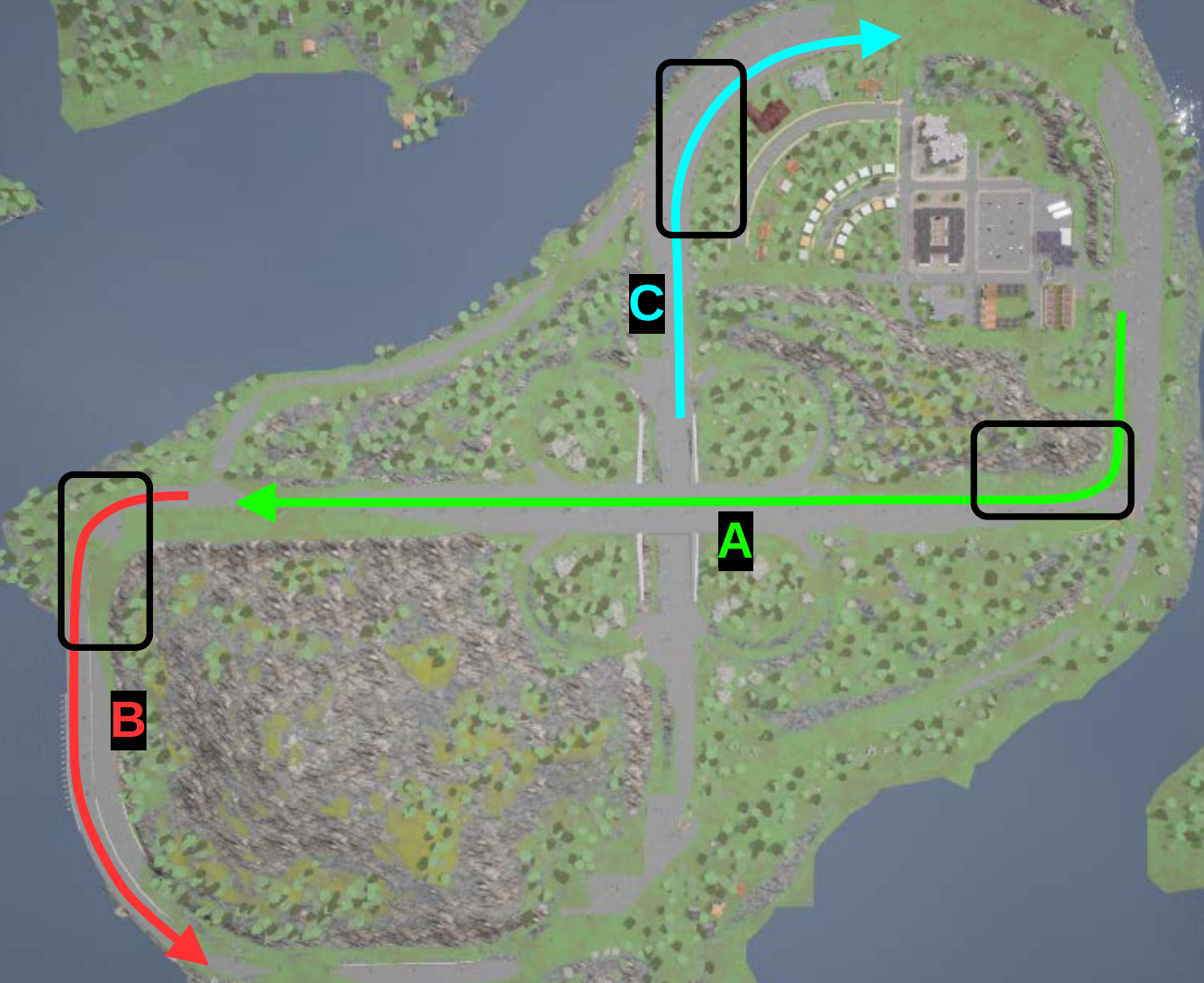}
\caption{Predefined evaluation routes in CARLA Town04. 
Route A emphasizes short straight segment followed by 90$^\circ$ right turn then a long straight segment.
Route B starts with a 90$^\circ$ left turn contains followed by a short straight segment and a sustained left-hand curvature. 
Route C contains short straight $\rightarrow$ and right-hand curvature.
Colored arrows indicate nominal driving direction. 
Key subsets (black boxes) isolate steering-intensive segments used for focused stability analysis.}
\label{fig:routes}
\end{figure}

Experiments are performed in CARLA Town04, which contains straight segments, 90$^\circ$ turns, and sustained curvature. Three predefined routes (A, B, and C) are constructed (Fig.~\ref{fig:routes}) to expose distinct lateral-control demands.
\textit{Route A:} short straight $\rightarrow$ 90$^\circ$ right turn $\rightarrow$ long straight. 
\textit{Route B:} 90$^\circ$ left turn $\rightarrow$ short straight $\rightarrow$ sustained left-hand curvature. 
\textit{Route C:} short straight $\rightarrow$ sustained right-hand curvature.

For each route, we additionally analyze a corresponding \emph{key} subset that isolates steering-intensive segments where latency-induced instability first emerges, enabling focused curvature-driven evaluation.

\subsection{Controller Configuration}
\label{sec:exp_controller}

All experiments use the deterministic client-side controller described in Section~\ref{sec:method_autonomy}. Lane boundaries are estimated from the image stream and converted to a BEV centerline. Lateral control uses speed-adaptive Pure Pursuit, and longitudinal motion is regulated by a PI controller targeting 30\,km/h. Controller parameters remain fixed across latency conditions.

Baseline runs (L0) include only inherent transport and codec latency and establish reference tracking statistics for comparison.

\subsection{Latency Injection Protocol}
\label{sec:exp_latency}

Latency is injected using \texttt{tc netem}, applied independently to the video (server$\rightarrow$client) and control (client$\rightarrow$server) channels to manipulate perception delay ($\tau_v$) and actuation delay ($\tau_c$).

The evaluated conditions are listed in Table~\ref{tab:latency_conditions}. Jitter and packet loss are excluded to isolate deterministic delay effects.
For each condition, 30 runs (6 routes × 5 repetitions) are performed. Across L0–L5, this yields 180 closed-loop runs.

\begin{table}[h]
\centering
\caption{Latency Conditions}
\label{tab:latency_conditions}
\begin{tabular}{c|c|c}
\hline
Condition & Video Delay ($\tau_v$) & Control Delay ($\tau_c$) \\
\hline
L0 & 0 ms   & 0 ms   \\
L1 & 75 ms & 0 ms   \\
L2 & 150 ms & 0 ms   \\
L3 & 225 ms & 0 ms   \\
L4 & 150 ms & 75 ms \\
L5 & 225 ms & 100 ms \\
\hline
\end{tabular}
\end{table}

\subsection{Measurement and Logging}
\label{sec:exp_logging}

Each run records one-way video and control latency ($\tau_v$, $\tau_c$), lateral and heading errors, steering statistics, and route completion status.

Latency measurements incorporate Chrony-based clock alignment (Section~\ref{sec:method_timesync}). Injected delay is verified by comparing measured one-way latency distributions before and after NetEm configuration.

\subsection{Evaluation Metrics}
\label{sec:exp_metrics}

Performance is quantified using mean absolute lateral error (MAE), root-mean-square error (RMS), and 95th-percentile absolute cross-track error. Stability is further assessed via maximum lateral deviation and route completion rate.
We additionally report collision rate (Coll.) and lane-invasion events (LaneInv) as discrete safety indicators derived from simulator event flags.

Results under latency condition $L_i$ are compared to baseline $L_0$ to characterize degradation trends. Environmental conditions remain fixed, and the controller is deterministic to ensure that performance variations arise solely from latency manipulation.

    \section{Results}
\label{sec:results}

This section presents quantitative and qualitative evaluation of closed-loop lane-keeping performance under increasing latency (L0–L5), aggregated over 30 runs per condition.

\subsection{Latency Verification}
\label{subsec:latency-verification}
\begin{figure}[!t]
\centering
\includegraphics[width=\linewidth]{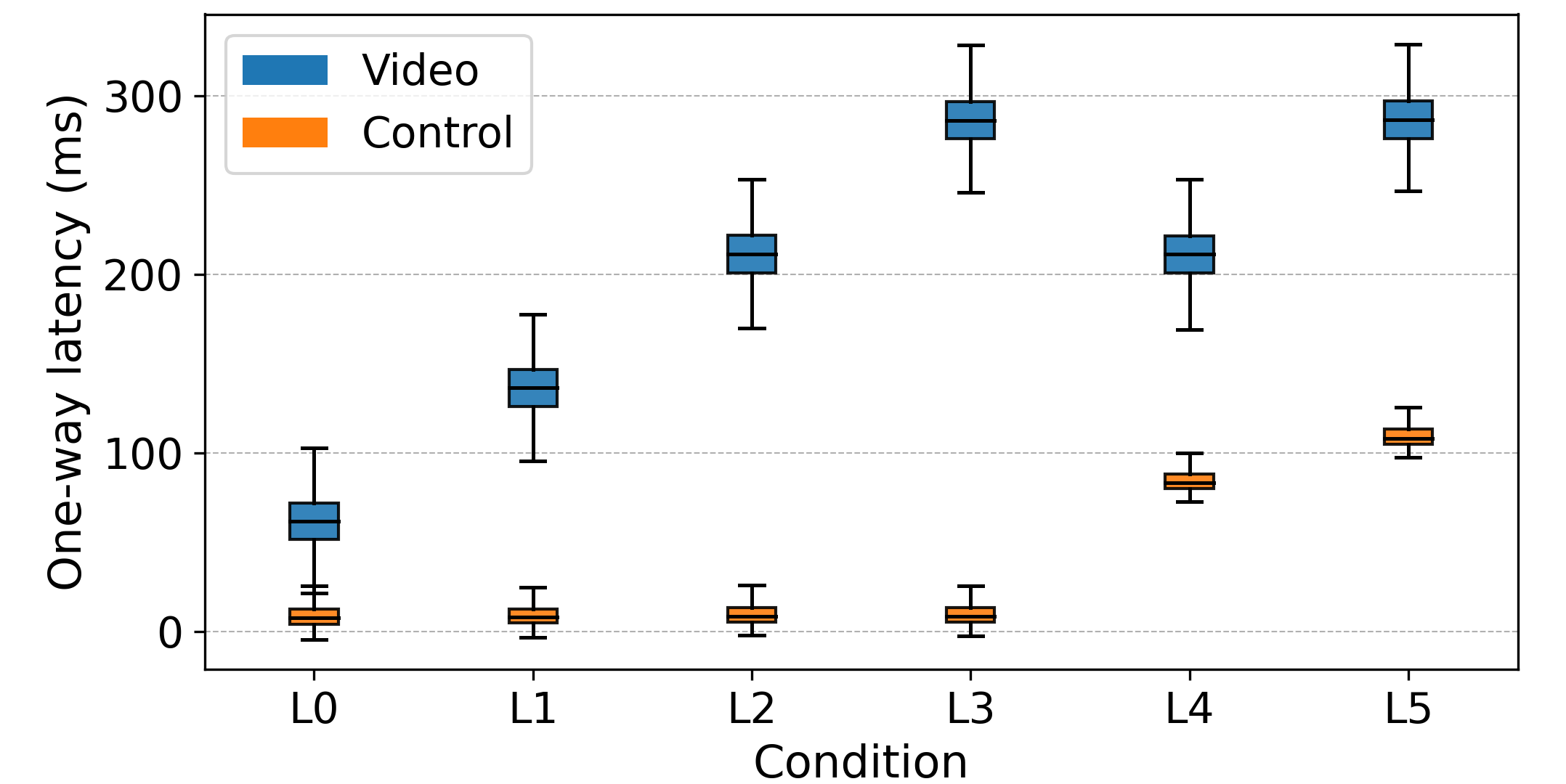}
\caption{Verification of injected one-way latencies across all routes (Town04) for L0--L5. 
For each condition, colored boxplots show video latency ($\tau_v$) and control-command latency ($\tau_c$) on the same axis.
L0--L3 increase video latency only.
L4 and L5 reuse the video-delay profiles of L2 and L3, respectively, while injecting additional control-command latency (visible as a distinct $\tau_c$ distribution at L4–L5).
}
\label{fig:latency_boxplot}
\end{figure}

Before analyzing control performance, we first verify that the measured end-to-end latencies match the intended experimental configurations and that the baseline teleoperation stack operates under low-delay conditions.

Figure~\ref{fig:latency_boxplot} reports the measured one-way latencies for both the video stream ($\tau_v$) and the control-command channel ($\tau_c$) across all routes under conditions L0--L5.
\begin{figure}[!t]
\centering
\includegraphics[width=\linewidth]{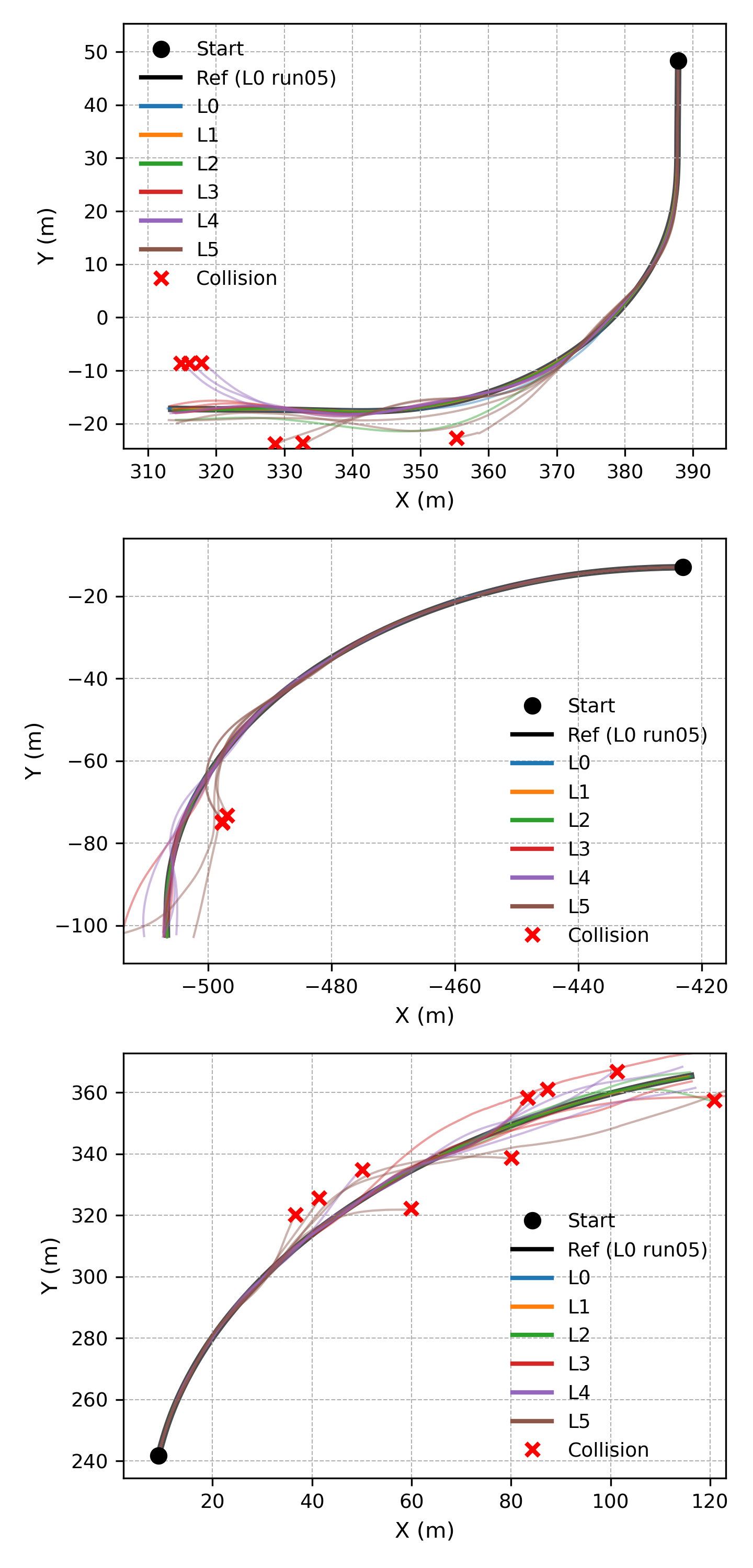}
\caption{Trajectory overlays for key route segments (Town04) under latency conditions L0–L5. Top: Route A Key, Middle: Route B Key, Bottom: Route C Key. Thin colored lines show individual runs; the thick black curve denotes the reference trajectory (L0 run05). Red crosses indicate aborted runs due to collision. Increasing perception latency induces growing deviation and phase-lag effects, while additional control-command delay (L4–L5) further amplifies instability and termination frequency, demonstrating nonlinear degradation of closed-loop stability.}
\label{fig:traj_key}
\end{figure}

Under the baseline condition (L0), the LAVT system exhibits low inherent network delay over WiFi~5, with a median video latency of approximately 62\,ms (p95 $\approx$ 86\,ms) and a median control-command latency of approximately 8--10\,ms. These values reflect encoding, transport, buffering, and wireless transmission overhead, establishing that the system operates in a stable low-latency regime prior to artificial delay injection.

Conditions L1--L3 increase video latency only. The measured video medians shift deterministically to approximately 136\,ms (L1), 211\,ms (L2), and 287\,ms (L3), with tight interquartile ranges across all routes. This confirms that NetEm successfully imposed controlled perception delay without introducing significant jitter.

Conditions L4 and L5 are designed to isolate the effect of \emph{control-command} delay. In these configurations, the video-delay profiles of L2 and L3 are reused, respectively, while additional latency is injected exclusively in the control channel. The measurements reflect this design: video latency at L4 closely matches L2, and at L5 matches L3, whereas control-command latency increases sharply (median $\approx$ 83\,ms at L4 and $\approx$ 108\,ms at L5) compared to the $\approx$8--10\,ms observed for L0--L3.

The narrow distributions and close agreement between configured and measured values confirm deterministic delay injection and accurate clock alignment.

\subsection{Trajectory Behavior Under Increasing Latency}

Figure~\ref{fig:traj_key} visualizes representative trajectory overlays for key routes under all latency conditions. Each colored trace corresponds to one run, while the black curve denotes the reference trajectory (L0 run05). Red markers indicate termination due to collision events.

Under L0, trajectories remain tightly clustered around the reference, indicating stable closed-loop behavior. Mild perception delay (L1) introduces small deviations, particularly in curvature transitions.

At L2 ($\tau_v \approx 211$\,ms), dispersion increases substantially and multiple runs fail, with instability emerging first in curved segments. At L3, oscillatory steering and overshoot produce frequent lane departures.

L4 shows slightly improved geometry among surviving runs due to survivorship bias, while L5 exhibits widespread early termination and large-amplitude oscillations consistent with delayed-feedback instability.

\begin{figure}[!t]
\centering
\includegraphics[width=\linewidth]{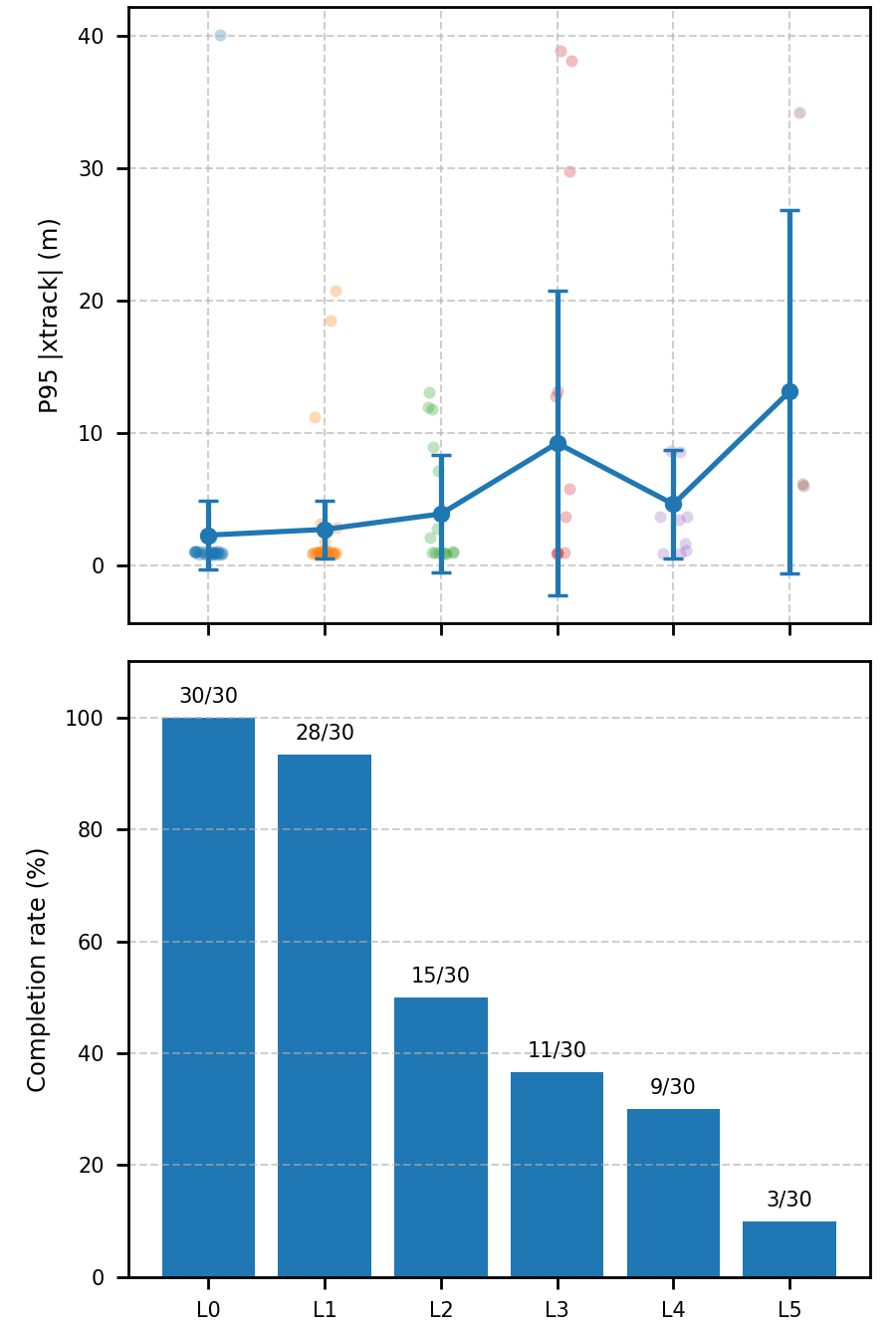}
\caption{Latency-induced degradation in closed-loop lane keeping (Town04; routes A–C and key subsets). 
Top: Route-balanced mean 95th-percentile absolute cross-track error computed over runs that successfully complete the route without collision; faint markers indicate per-run values. Error bars denote 95\% confidence intervals of the mean. 
Bottom: Route completion rate over all runs; annotations show completed/total runs (out of 30). 
Tracking degradation emerges at L2 ($\approx$ 211 ms one-way perception delay), while completion exhibits a nonlinear collapse beyond this threshold. L4 isolates additional control-command delay under matched perception latency.}
\label{fig:degradation}
\end{figure}

\subsection{Quantitative Degradation Trends}

Figure~\ref{fig:degradation} summarizes performance degradation as a function of latency using two complementary metrics: (i) the 95th-percentile absolute cross-track error computed over completed runs only, and (ii) the route completion rate computed over all runs.
Tracking error is computed exclusively over runs that successfully complete the route without collision, thereby avoiding artificial inflation of error statistics due to premature termination.

\paragraph{Tracking Performance.}

Under L0, P95 cross-track error is approximately 2.3\,m. Mild delay (L1) produces modest degradation. At L2, error nearly doubles while completion drops to 50\%, indicating onset of instability. At L3, oscillatory corrections further increase error. The apparent improvement at L4 reflects survivorship bias, as unstable runs terminate early. At L5, P95 exceeds 13\,m among surviving runs while completion collapses to approximately 10\%.

\paragraph{Completion Collapse.}

Completion rate exhibits nonlinear degradation: robust at L0–L1, partial failure at L2, and rapid collapse beyond L3. Instability emerges between 150 ms and 225 ms one-way perception latency, where completion falls below 50\%. Notably, completion degradation precedes extreme geometric divergence, indicating loss of closed-loop stability before catastrophic tracking error appears in surviving runs.

Table~\ref{tab:main_performance_latency} summarizes aggregate driving performance metrics across all routes for each latency condition, including completion rate, collision frequency, lane-invasion events, and 95th-percentile cross-track error.
Consistent with Fig.~\ref{fig:degradation}, the aggregate statistics confirm nonlinear degradation of closed-loop lane keeping as perception latency increases, with a sharp transition between 150 ms and 225 ms one-way delay. The system remains stable at 75 ms, exhibits clear instability by 150 ms, and undergoes widespread collapse by 225 ms. Additional control-command latency further accelerates this collapse.

\begin{table}[t]
\centering
\caption{Driving performance under latency conditions (Town04, all routes).}
\label{tab:main_performance_latency}
\setlength{\tabcolsep}{10pt}
\renewcommand{\arraystretch}{1.1}
\begin{tabular}{c c c c c}
\hline
\textbf{Cond.} & \textbf{Comp. (\%)} & \textbf{Coll.} & \textbf{LaneInv} & \textbf{$P95_{x}$ (m)} \\
\hline
L0 & 100 & 0.20 & 0.40 & 2.30 \\
L1 & 93.3 & 0.23 & 3.97 & 2.73 \\
L2 & 50.0 & 0.67 & 15.10 & 3.91 \\
L3 & 36.7 & 0.83 & 22.13 & 9.25 \\
L4 & 30.0 & 0.77 & 18.80 & 4.64 \\
L5 & 10.0 & 0.97 & 16.83 & 13.15 \\
\hline
\end{tabular}
\end{table}

    \section{Discussion}
\label{sec:discussion}

\subsection{Delay-Induced Instability Mechanisms}

The observed degradation is consistent with classical control theory. Perception delay introduces phase lag, causing corrective steering to act on stale lane geometry. As $\tau_v$ increases, overshoot and counter-corrections produce oscillatory behavior. The sharp transition between L2 and L3 suggests a stability boundary rather than gradual decay.

The divergence between tracking error and completion rate highlights survivorship bias: unstable runs terminate early, leaving only comparatively stable trajectories in completed-run statistics. Reporting both metrics is therefore essential for accurate stability assessment.

\subsection{Implications for Teleoperation}

Perception delay in the 150–300\,ms range substantially reduces stability margins in vision-driven teleoperation. As shown in Table II, completion drops from 100\% at L0 to 50\% at L2 and 10\% at L5, while P95 cross-track error increases nearly sixfold between L0 and L5. Network variability may intermittently push systems beyond this boundary unless mitigation mechanisms are applied.

Although compensation strategies are not evaluated here, the identified stability threshold provides a baseline for assessing predictive perception or latency-compensation methods.
    \section{Conclusion and Future Work}

This paper presented a systematic experimental characterization of network-induced latency in vision-based teleoperation. Using controlled delay injection and 180 closed-loop experiments across diverse geometric regimes, we identified an empirical stability transition region between 150 ms and 225 ms one-way perception delay. Below this threshold, lane keeping remains robust; beyond it, completion rate collapses rapidly. Additional control-command delay further accelerates instability even when visual delay is unchanged.

These findings demonstrate that perception-driven teleoperation exhibits sharp delay-dependent stability limits rather than gradual degradation. Quantifying this boundary is essential for safe deployment and for benchmarking latency-compensation strategies.

\paragraph{Limitations.}
The study is conducted in simulation under deterministic delay without jitter or packet loss. Dynamic traffic, weather variation, and higher-speed driving conditions are not considered. Furthermore, the study evaluates a deterministic vision-based lane-keeping controller to isolate latency effects; stability boundaries may shift under alternative control architectures. 

\paragraph{Future Work.}
Future work will extend this characterization to real-vehicle experiments and stochastic network conditions, including jitter and packet loss. 
Additional controller architectures will be evaluated across several road types in different towns. 
In addition, predictive perception and latency-compensation techniques will be evaluated against the baseline stability threshold identified in this study.

    \bibliographystyle{unsrt}
\bibliography{3_end/references.bib}
\end{document}